# Cognitive Multi-Agent Search and Rescue: Language-Grounded Hierarchical Reinforcement Learning for Collaborative Robotics

Sepideh Ravanbakhsh[1,*] Hamidreza Moradi[2]

[1]Computer Engineering Department, Science and Research Branch, Islamic Azad University (SRBIAU), Tehran, Iran

2Department of Mechanical Engineering and Engineering Science, The University of North Carolina at Charlotte, Charlotte, North Carolina, USA

[*]Corresponding author: ravanbakhsh80@gmail.com; hmoradi@charlotte.edu

## Abstract

Effective search and rescue (SAR) operations require autonomous systems capable of coordinating multiple heterogeneous agents, adapting to dynamic hazards, and aligning with human mission intent. Existing language-guided reinforcement learning (RL) methods primarily focus on single-agent instruction following and lack mechanisms for team-level coordination in complex, partially observable environments. This paper presents a Cognitive Multi-Agent Framework based on Language-Grounded Hierarchical Reinforcement Learning (LG-HRL) that integrates natural-language understanding, hierarchical policy decomposition, and a shared cognitive workspace to achieve cohesive multi-agent behavior in SAR scenarios. Human instructions are interpreted by a large language model (LLM) and transformed into structured semantic cues that guide high-level strategy selection and role assignment across UAV and UGV agents. A formal Language-Grounded Multi-Agent Hierarchical Decision Process (LG-MAHDP) is introduced to model the interplay between semantic grounding, hierarchical control, and distributed coordination. Comprehensive experiments in a simulated disaster environment show that LG-HRL improves learning speed by 2.3×, increases coordination coherence by 37%, and enhances subgoal accuracy and hazard avoidance by 46% and 31%, respectively. The system achieves 20–30% higher victim rescue rates, demonstrates emergent role specialization, and adapts immediately to human verbal updates. Ablation studies confirm that both the shared cognitive workspace and language grounding are critical to overall system performance. Overall, this work establishes a new paradigm for language-guided multi-agent SAR robotics, demonstrating that semantic grounding combined with hierarchical reasoning enables safer, more efficient, and more interpretable autonomous rescue operations.

## 1. Introduction

Natural disasters such as earthquakes, floods, wildfires, and industrial accidents continue to pose significant threats to human life and infrastructure. In these high-risk situations, timely access to reliable information and rapid deployment of coordinated rescue operations often determine mission success. Over the past decade, autonomous robots—ranging from unmanned aerial vehicles (UAVs) to ground-based rescue platforms—have begun to play an increasingly central role in assisting first responders by surveying hazardous regions, identifying survivors, and supporting logistics in areas inaccessible to humans [1]. Their ability to enter unstable structures, traverse difficult terrain, and operate continuously makes them invaluable in disaster response.

Despite these advantages, traditional SAR robotics remains constrained by single-agent autonomy, hand-engineered control pipelines, and rule-based decision mechanisms. These systems struggle with the complexity and unpredictability of real disaster environments, where conditions evolve rapidly and ambiguity is common. More critically, existing systems lack cognitive capabilities: they cannot infer intent, reason about high-level missions, adjust strategies based on newly acquired information, or coordinate seamlessly with other agents [2].

Recent developments in artificial intelligence—particularly Large Language Models (LLMs) and Hierarchical Reinforcement Learning (HRL)—signal a transformative opportunity for SAR robotics. LLMs provide a means for robots to interpret natural-language instructions, extract semantic meaning, and ground human intent in operational tasks. HRL, on the other hand, offers a structured framework for decomposing complex missions into manageable subtasks, enabling more stable learning over long horizons. When combined within a multi-agent context, these advances open the door to collaborative, adaptive, and language-informed SAR systems capable of functioning with a level of competence and flexibility far beyond traditional approaches [3,4].

Integrating advanced AI capabilities into real-world search and rescue (SAR) operations remains a significant challenge. Multi-agent systems must sustain coherent team behavior despite uncertainty, environmental variability, communication delays, and partial observability. Effective SAR missions further require that robots share information efficiently, dynamically allocate roles, and adapt their actions as new situational cues emerge. At the same time, human supervisors frequently express mission guidance through natural language—describing priorities, hazards, or suspected victim locations—which conventional robotic systems are unable to interpret or convert into actionable commands [5]. For autonomous agents to function as reliable partners rather than isolated tools, they must be equipped with mechanisms to understand and contextualize human instructions, translate semantic information into hierarchical task structures, coordinate actions with teammates in real time, and adjust decision-making processes as exploration unfolds [4,5]. These challenges motivate the central research question of this work: How can natural language understanding be fused with hierarchical reinforcement learning to achieve scalable, cooperative, and situationally aware behavior in multi-agent search and rescue robotics? Addressing this question targets a critical gap in current SAR research, where most language-guided frameworks focus on single-agent exploration and lack mechanisms for collaborative, cognitively integrated multi-robot teams.

The overarching goal of this study is to develop a Language-Grounded Hierarchical Reinforcement Learning (LG-HRL) framework that enables cognitively coordinated multi-agent operation in SAR environments. The proposed approach utilizes the semantic reasoning capabilities of large language models (LLMs) to interpret natural language mission directives and align them with hierarchical reinforcement learning policies optimized for multi-agent collaboration. Toward this goal, the study aims to: (1) design a unified cognitive architecture that integrates language grounding with HRL-based policy hierarchies; (2) empower heterogeneous agents—such as UAVs and ground robots—to autonomously convert high-level human intent into coordinated low-level actions; (3) enhance team adaptability and resilience by implementing a shared cognitive workspace for real-time information exchange, task redistribution, and cooperative planning; (4) strengthen human–AI interaction and interpretability through language-based explanations, enabling dynamic mission adjustment; and (5) validate the system's effectiveness in simulated disaster environments characterized by sparse rewards, long planning horizons, and operational uncertainty. Collectively, these objectives advance the development of a collaborative, explainable, and robust multi-agent SAR system capable of functioning reliably under the demanding conditions of disaster response.

## 2. Literature Review

### 2.1 Applications of Artificial Intelligence and Machine Learning Across Domains

Artificial intelligence (AI) and machine learning (ML) have emerged as foundational technologies across a wide spectrum of scientific, engineering, and socio-technical domains, enabling intelligent perception, adaptive decision-making, and autonomous coordination in complex environments. Recent advances in deep learning, explainable AI, graph-based modeling, and data-driven optimization have significantly expanded the applicability of intelligent systems beyond narrowly defined tasks toward integrated, multi-agent, and context-aware frameworks.

In healthcare and medical diagnostics, deep learning has demonstrated strong capabilities in perception-driven intelligence, real-time inference, and decision support. Fusion-based models combining deep neural networks with explainable AI and rule-based reasoning have been shown to improve robustness and interpretability in brain tumor classification, addressing trust and transparency requirements in safety-critical systems [6]. Similarly, deep learning frameworks designed for real-time medical image processing have enhanced diagnostic accuracy while meeting stringent latency constraints, highlighting the role of efficient perception pipelines in time-sensitive environments [7]. Complementary work on intelligent calibration and parameter control has applied data-driven analysis to improve the reliability of X-ray diffractometry and electron-dispersive spectroscopy, demonstrating how ML-informed optimization supports high-fidelity sensing and instrumentation [8], [9]. These studies collectively underscore the importance of combining learning-based perception with explainability, domain constraints, and system-level reliability. Beyond medical imaging, AI has been widely adopted in natural language processing (NLP) and information systems, which are essential components of language-grounded and human–machine collaborative intelligence. Hierarchical transformer-based architectures have achieved notable success in Persian text readability assessment, illustrating the adaptability of modern language models to linguistically complex and low-resource contexts [10]. Earlier work on service-oriented architectures for secure data hiding reflects the evolution from static information systems toward intelligent, distributed, and learning-enabled infrastructures [1]. Such advances in language understanding and secure information processing are directly relevant to multi-agent systems that rely on natural language grounding, communication, and coordination.

AI and ML have also been extensively applied in social systems, education, and behavioral analytics. Data-driven competency frameworks have been proposed to model entrepreneurial skill development within sustainable education systems, highlighting the role of intelligent modeling in complex human-centered domains [12]. In parallel, systematic reviews of e-sports research and physical education analytics demonstrate the growing influence of AI-driven evaluation, performance analysis, and digital interaction modeling in emerging cyber–physical learning environments [13], [14]. These studies illustrate how intelligent systems increasingly operate at the intersection of human behavior, digital platforms, and adaptive feedback.

In urban analytics and smart city applications, ML has been leveraged to model complex spatial, social, and infrastructural interactions. Multivariate learning approaches have been used to analyze the relationship between urban street network configurations and property crime, revealing non-linear spatial dependencies that are difficult to capture with classical models [15]. Intelligent forecasting methods based on fuzzy logic have also been applied to traffic density prediction, supporting adaptive control and decision-making in dynamic transportation networks [16]. Such urban-scale intelligent systems parallel the challenges faced in multi-agent search and rescue scenarios, where agents must reason over spatial structure, uncertainty, and dynamic environmental feedback. Machine learning has further played a critical role in performance modeling, optimization, and intelligent infrastructure. At the systems level, graph-structured representations have proven effective for reasoning about execution dynamics and resource utilization in complex computing platforms. Performance-aware graph models that explicitly account for hardware heterogeneity and scheduling constraints have been proposed to optimize parallel code execution, illustrating how structural knowledge of computation can inform adaptive decision-making [17]. This perspective closely parallels multi-agent robotic systems, where agents must coordinate actions while respecting communication, timing, and resource limitations.

Interdisciplinary applications of AI and ML have expanded into materials science, nanotechnology, and biomedical engineering, where data-driven modeling supports design optimization and system intelligence. From a modeling perspective, mesoscale approaches based on representative volume elements provide an effective means of capturing heterogeneity and interaction effects in complex physical systems. This is exemplified by studies on the mechanical properties of concrete with aggregates and voids, where mesoscale RVEs are used to bridge local structural features and macroscopic behavior [18]. Learning-assisted analysis has been applied to nanocomposite membranes for enhanced gas separation [19], pharmaceutical formulation optimization [20,21], drug delivery systems [22], and theranostic nanoplatforms for cancer treatment [23-25]. Additional studies have explored dual-trigger release mechanisms [26], ultrasound-guided chemoradiotherapy [27], targeted nanodroplet enhancement of radiotherapy [28], and dual-modal imaging systems [29]. Intelligent materials and bioactive scaffold design have also benefited from data-driven approaches in tissue engineering and biomaterials development [30-36]. Although primarily experimental, these works increasingly rely on multivariate analysis, optimization, and learning-informed decision support, reflecting the growing convergence of AI with physical system design. Finally, AI-driven modeling and optimization have been applied in energy harvesting and sensing technologies, particularly in infrared nanoantenna systems. Intelligent design principles have enabled diode-less rectification using planar cross-bowtie nanoantenna arrays [37], multiband plasmonic nanoantenna structures [38], field-emission-based infrared rectification mechanisms [39], and nanoantenna arrays optimized for mid-infrared energy harvesting [40]. These studies highlight how learning-informed design supports adaptive performance in dynamic electromagnetic environments, a theme shared with autonomous robotic systems operating under uncertain sensing and energy constraints.

Collectively, these applications demonstrate that modern AI and ML systems emphasize adaptability, interpretability, scalability, and cross-domain integration. Across perception, language understanding, optimization, and physical system design, intelligent models increasingly operate in environments characterized by uncertainty, dynamic interactions, and distributed agents. These characteristics directly motivate the development of cognitive multi-agent frameworks, such as language-grounded hierarchical reinforcement learning, for collaborative robotic search and rescue operations in complex real-world environments.

### 2.2 Search and Rescue Robotics

Robotic systems have long been recognized as valuable assets in disaster response due to their ability to access environments too dangerous or unstable for human rescuers. Early generations of SAR robots relied heavily on teleoperation, where human operators manually navigated robots through collapsed structures or hazardous terrain. Although effective in certain conditions, teleoperation suffers from limited situational awareness, operator fatigue, and communication delays that hinder responsiveness. As autonomous navigation matured, research shifted toward exploration strategies such as frontier-based exploration, occupancy grid mapping, SLAM techniques, and probabilistic search algorithms designed to maximize coverage and reduce redundancy. UAVs became indispensable for aerial reconnaissance, providing rapid global views of disaster zones, while tracked or wheeled ground robots were deployed for close-proximity inspection and physical interaction with debris [41].

More advanced SAR systems introduced multi-robot task allocation, bio-inspired swarm strategies, and cooperative mapping to improve robustness. Nevertheless, many SAR deployments still rely on rule-based behaviors and manually tuned heuristics, which limit adaptability when environments evolve unpredictably—such as shifting rubble, blocked passages, or emergent hazards. Furthermore, most existing systems cannot ingest or reason about semantic-level information (e.g., “victims may be near staircases” or “avoid areas with chemical leakage”) expressed by human responders. This gap highlights an emerging need for cognitively enriched

SAR robotics capable of interpreting high-level human reasoning, dynamically adjusting mission priorities, and coordinating across heterogeneous platforms in a scalable manner [42].

### 2.3 Multi-Agent Reinforcement Learning (MARL)

Multi-Agent Reinforcement Learning (MARL) addresses decision-making in environments where multiple agents interact simultaneously, often with shared goals. Foundational methods such as independent Q-learning, centralized training and decentralized execution (CTDE), and various actor–critic architectures aim to mitigate the non-stationarity that arises when multiple learners update policies concurrently [43]. More recent advancements incorporate value factorization techniques (e.g., QMIX, VDN), policy-sharing strategies, and graph-based models that capture relational dependencies among agents. These approaches have enabled progress in tasks ranging from cooperative navigation and multi-robot coverage to adversarial pursuit–evasion scenarios [44].

Communication-enhanced MARL frameworks further introduce implicit or explicit communication channels, allowing agents to exchange latent states, belief models, or intentions. Mechanisms such as attention networks or learned communication protocols improve coordination reliability, especially under partial observability. Despite these advancements, significant challenges persist: MARL systems often struggle with scalability, reward sparsity, credit assignment, and coordination under dynamic conditions—all of which are central to SAR missions. Moreover, existing MARL research is largely disconnected from **human-centric guidance**; agents rarely incorporate instructions or semantic cues provided by operators, limiting the applicability of MARL to real-world rescue contexts where human oversight and domain expertise are indispensable [45].

### 2.4 Hierarchical Reinforcement Learning (HRL)

Hierarchical Reinforcement Learning provides a principled way to address long-horizon tasks by decomposing them into subgoals or temporally extended actions. The Options framework, Semi-MDP models, h-DQN, HIRO, and FeUdal Networks exemplify approaches that separate decision-making across multiple temporal layers. High-level controllers determine abstract goals or subgoals, while low-level controllers optimize continuous control or navigation policies. HRL is especially suited to domains, such as SAR, where tasks naturally consist of sequential stages: exploring unknown terrain, identifying potential victims, assessing structural stability, or delivering supplies [46].

HRL offers benefits including improved sample efficiency, structured learning, and the ability to handle environments with sparse or delayed rewards. However, classical HRL relies heavily on manually designed hierarchies, fixed subgoal templates, or engineered reward shaping. This limits its flexibility in real missions where objectives may shift dynamically based on operator input or situational developments. Furthermore, HRL methods typically lack semantic grounding: high-level policies cannot interpret natural language descriptions to automatically update subgoals or adjust hierarchy structure. Bridging HRL with human-intent interpretation is therefore a crucial step toward creating adaptable, real-world autonomous systems [45].

### 2.5 Language-Guided and Language-Grounded Learning

The integration of natural language understanding into robotic control has seen rapid progress in recent years, driven by advances in transformers, large language models (LLMs), and multimodal architectures. Early approaches relied on symbolic parsing or template-based instruction following, enabling robots to execute simple tasks specified in natural language. Later works in language-conditioned RL demonstrated that natural language can influence policy shaping, reward modulation, and exploration strategies. Systems such as SayCan, Inner Monologue, and grounded decision transformers illustrate how language models can propose or refine contextual subgoals, assist with planning, or explain action choices [47].

However, most existing language-guided robotic frameworks remain limited to single-agent tasks, controlled indoor environments, or constrained action spaces. Additionally, while LLMs excel at semantic reasoning, they typically lack mechanisms to directly interface with hierarchical decision policies or to coordinate behavior across multiple agents. Methods combining natural language with reinforcement learning often lack a cognitive substrate that allows robots to share reasoning, synchronize mission understanding, or collaboratively adapt based on human-provided instructions. As a result, language grounding in multi-agent, high-stakes environments such as SAR is still largely unexplored and presents a major frontier in robotics research.

### 2.6 Summary and Research Gap

Across the domains of SAR robotics, MARL, HRL, and language-guided learning, the literature demonstrates significant advancements yet remains fragmented in key ways. Traditional SAR systems lack the cognitive and collaborative depth needed to manage complex, evolving disaster environments. MARL provides coordination capabilities but seldom incorporates human-intent understanding. HRL offers temporal abstraction but does not naturally integrate semantic reasoning or dynamically generated subgoals. Language-guided approaches enable natural interaction with robots but are overwhelmingly tailored to single-agent setups and lack mechanisms for coordinating multi-agent teams [4, 5, 42].

These limitations collectively reveal a clear and critical research gap: there is no unified framework that integrates natural language understanding, hierarchical reinforcement learning, and multi-agent coordination into a cognitively coherent system for real-world search and rescue operations. Existing approaches fail to combine semantic interpretation, hierarchical task decomposition, and team-level cooperation in a manner suitable for dynamic and uncertain disaster environments. The present study addresses this gap by proposing a Language-Grounded Hierarchical Reinforcement Learning (LG-HRL) framework that tightly couples LLM-based semantic reasoning with hierarchical policy formation and distributed multi-agent coordination. By leveraging shared cognitive workspaces, dynamic subgoal formation, and natural language-driven task alignment, this framework establishes a foundation for scalable, interpretable, and human-aligned multi-robot SAR systems capable of operating under long-horizon, sparse-reward, and high-uncertainty conditions.

# 3. Methodology and System Architecture

## 3.1 Overview of the Cognitive Multi-Agent Framework

The proposed framework introduces a cognitively grounded multi-agent architecture for autonomous search and rescue (SAR) operations in complex and dynamic environments. Unlike traditional single-agent HRL systems, the framework enables a heterogeneous team of aerial and ground robots to operate collaboratively under human guidance while maintaining a synchronized understanding of mission-level objectives. The architecture consists of four integrated layers: a natural-language operator interface, a semantic interpretation module powered by a large language model (LLM), a cognitive coordination layer that aggregates shared knowledge, and a multi-agent execution layer implementing hierarchical reinforcement learning (HRL). As illustrated in Fig. 1**,** natural-language mission directives are first processed by the LLM-based Context Extractor, which transforms verbal instructions into structured contextual features. These semantic cues are merged with observations from the environment and priority information from the Information Space within a shared cognitive workspace, enabling the Strategic Decision Engine to generate refined high-level strategies for each agent. Worker modules then execute these strategies through low-level control policies, interacting with the environment and continuously updating the shared workspace. The key novelty of this framework lies in its ability to transform high-level human intent into structured representations that inform distributed, context-aware strategies across multiple robots. Through shared cognition, semantic grounding, and hierarchical decomposition, agents dynamically adjust their roles, allocate tasks cooperatively, and adapt to emerging environmental cues, maintaining coordinated performance throughout the SAR mission.

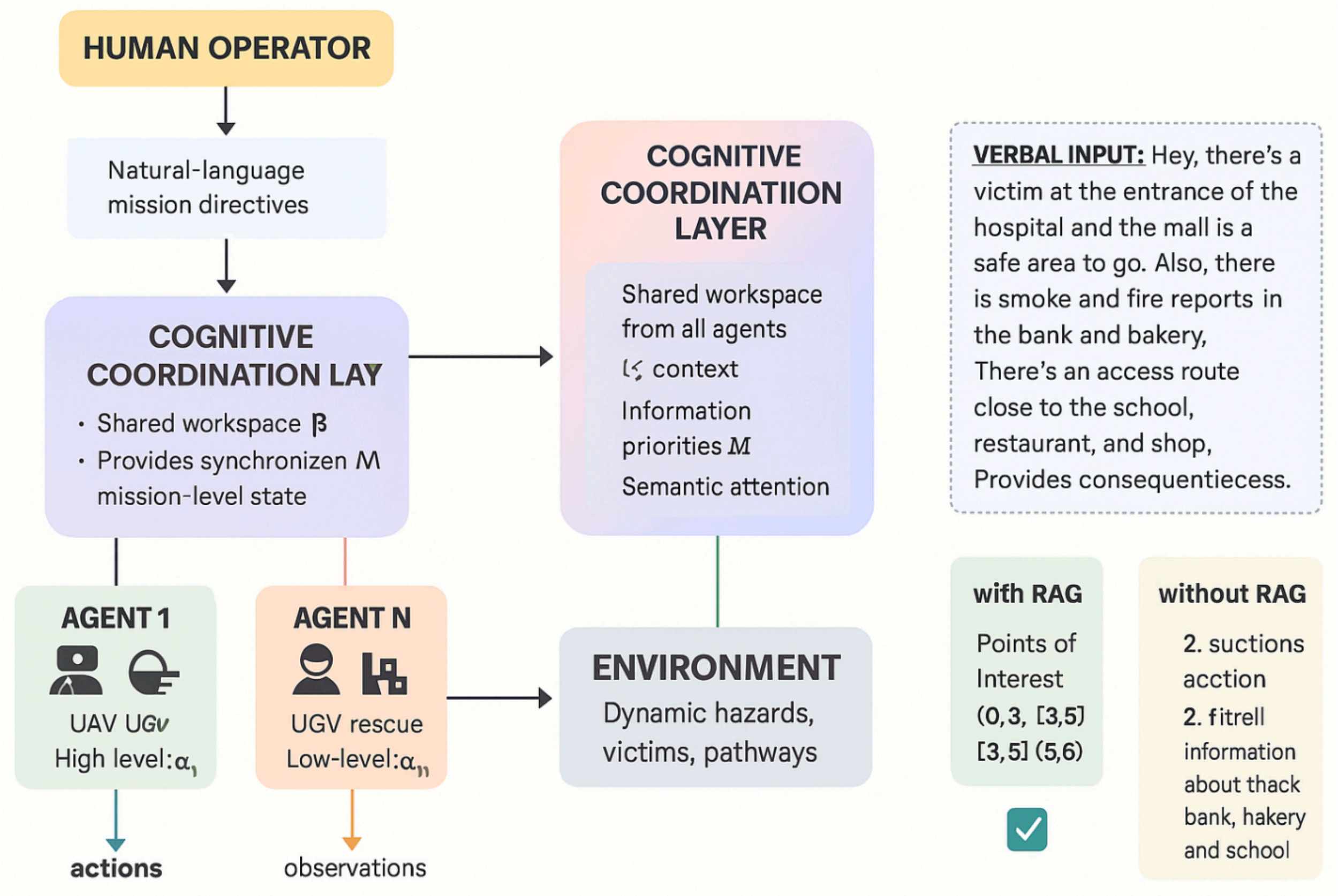

Fig. 1. Cognitive Multi-Agent Framework for Language-Grounded Hierarchical Decision-Making. *Left:* Verbal input $v$ from the human operator is processed by the LLM-driven Context Extractor to produce contextual features $C$. These features, together with observations $s$ from the Environment and information priorities $M$ from the Information Space, are integrated within the Strategic Decision Engine (SDE). The Attention Space refines strategy selection by mapping semantic cues and priority information via $\Psi(C, M)$. The SDE outputs high-level strategies $\omega$, which the Worker modules translate into low-level actions $\alpha$ executed within the Environment. The resulting observations are fed back into the shared workspace, enabling continuous learning and adaptation. *Right:* Comparison of LLM outputs with and without retrieval-augmented generation (RAG) in simulated SAR operations, showing that RAG enhances contextual precision, hazard annotation, and task-specific semantic structure in generated outputs.

### 3.2 Language-Grounded Hierarchical Reinforcement Learning (LG-HRL)

The LG-HRL module fuses high-level semantic reasoning with low-level policy execution. Human-issued natural language instructions are processed by the LLM, which extracts contextual features capturing mission priorities, environmental hazards, and task-level goals. These contextual features guide the high-level controller, which selects abstract strategies for the team such as reconnaissance, victim search, or hazard avoidance. Each chosen strategy activates a corresponding low-level sub-policy for each agent, enabling navigation, manipulation, sensing, and interaction behaviors. The hierarchical structure allows agents to operate over long temporal horizons while retaining the flexibility to revise their plans as contextual updates emerge. This

semantic–hierarchical fusion enables task decomposition to be dynamically reconfigured based on operator input, environmental evolution, and team performance [4, 47].

### 3.3 Modeling the Multi-Agent Semantic–Hierarchical Decision Process

To formalize the interaction between language, hierarchy, and multi-agent cooperation, we define a Language-Grounded Multi-Agent Hierarchical Decision Process (LG-MAHDP). The process consists of the tuple:

$$\mathcal{M} = (\mathcal{S}, \mathcal{A}_1, \dots, \mathcal{A}_N, \Omega, \beta_\omega, P, R, \gamma).$$

The global state space $\mathcal{S}$is maintained within a shared cognitive workspace that aggregates local observations, contextual cues, and mission-level information. Language inputs $V = \{v_1, \dots, v_m\}$are mapped into contextual features $C = \{c_1, \dots, c_n\}$using a transformation function $L: V \to C$parameterized by an LLM. An information-priority structure

$$M = \{(i_k, p_k)\},$$

expresses weighted mission constraints such as hazard-criticality, urgency of victim retrieval, and safe-route preferences.

High-level decisions are selected by a semantic policy

$$\pi_\Omega: \mathcal{S} \to \Omega,$$

which maps shared states to macro-strategies. Each agent executes a low-level sub-policy

$$\pi_{\omega,i}: \mathcal{S}_i \to \mathcal{A}_i,$$

optimized through reinforcement learning. Strategy and action values are computed as

$$Q(s, \omega) = \mathbb{E}[\sum_{t=0}^{T} \gamma^t R(s_t, \omega)], Q(s, \omega, a_i) = \mathbb{E}[\sum_{t=0}^{T} \gamma^t R(s_t, a_{i,t}) \mid \omega].$$

The shared cognitive workspace maintains the belief state

$$B_t = \{s_t^1, \dots, s_t^N, C_t, M_t, H_t\},$$

where $H_t$stores communication history. A semantic attention mechanism $\Psi(C, M)$refines the hierarchical policies by highlighting important contextual features. The joint multi-agent objective is to maximize

$$J = \mathbb{E}[\sum_{t=0}^{T} \gamma^t R(s_t, a_{1,t}, \dots, a_{N,t}) \mid \pi_\Omega, \{\pi_{\omega,i}\}, \Psi].$$

This formulation provides a principled mathematical foundation for semantic-driven hierarchical multi-agent control [43], distinguishing the system from classical HRL or purely language-guided single-agent frameworks.

### 3.4 Cognitive Coordination Layer

The cognitive workspace serves as the central substrate for distributed situational awareness and team-level decision-making. Agents continuously publish local observations, intermediate results, and hazard detections to the workspace, which aggregates these inputs into a coherent mission state. Semantic cues extracted from the LLM are grounded into this representation, enabling agents to reason about mission-wide constraints rather than acting solely on their local perceptions. Agents retrieve relevant updates from the workspace and autonomously adjust their behaviors, enabling coordinated coverage, reduced redundancy, and shared understanding of role assignments. This layer effectively replaces explicit agent-to-agent communication with a structured collective memory, improving scalability and robustness under limited bandwidth [49].

### 3.5 Human–AI Interaction Loop

The framework supports a continuous, bidirectional interaction loop between human supervisors and the robotic team. Operators provide high-level guidance using natural language, which the LLM interprets into structured semantic features. These features modify mission objectives, adjust priorities in the information space, and influence high-level strategy selection. Conversely, agents generate natural-language summaries describing discovered hazards, potential victims, and mission progress. This loop fosters interpretability and trust, allowing human operators to steer the system without micromanagement and enabling the multi-agent team to request clarification in ambiguous scenarios [50].

### 3.6 Algorithmic Workflow and Computational Considerations

The full operational pipeline progresses through iterative semantic interpretation, hierarchical strategy assignment, low-level policy execution, and workspace synchronization. The LLM inference frequency is modulated based on environmental novelty or explicit operator intervention, reducing computational load. Communication overhead is minimized through compact, prioritized state updates in the shared workspace. Hierarchical decision-making distributes complexity across temporal scales, reducing policy-space dimensionality. In combination, these mechanisms ensure that the system can operate on edge-deployed hardware typical of field robotics scenarios while maintaining real-time responsiveness.

## 4. Experiments

### 4.1 Experimental Objectives

The experimental evaluation examines the effectiveness of the proposed LG-HRL architecture in enabling autonomous multi-agent coordination grounded in natural-language input. Specifically, the experiments evaluate: (1) the consistency and accuracy with which agents operationalize operator-issued verbal instructions; (2) the degree of coordination achieved across heterogeneous agents; (3) improvements in learning efficiency enabled by hierarchical abstraction; (4) the influence of semantic grounding on hazard avoidance and victim rescue performance; and (5) the scalability of the framework under increasingly complex SAR environments [50].

### 4.2 Simulation Environment Setup

A multi-agent disaster simulation environment was constructed to emulate realistic SAR conditions involving uncertainty, partial observability, and stochastic hazards. The environment includes collapsed structures, blocked pathways, multiple hazard zones, and clusters of victims distributed across spatially diverse segments. UAVs and UGVs operate under different sensing modalities: UAVs provide broad but noisy top-down perception, while UGVs acquire detailed local observations. Human operators inject natural-language instructions during the episode, such as *"Prioritize victims near the collapsed factory; avoid the northern chemical spill; send drones ahead to map escape routes."* The LLM transforms these instructions into structured contextual features used to update high-level strategies in real time. The simulation is built in Python using an extended Gymnasium interface with support for multi-agent asynchronous execution and communication constraints.

### 4.3 Agent Team Configuration

Experiments were conducted using a team of four heterogeneous robots: two UAVs responsible for reconnaissance and hazard detection, and two UGVs responsible for navigation, victim assistance, and close-proximity exploration. Agents possess limited computing and sensing capabilities, reflecting deployment constraints typical in SAR missions. Each agent's action space includes primitive navigation, sensor activation, semantic communication updates, and specialized rescue actions. Communication is mediated through the cognitive workspace, with simulated bandwidth constraints.

### 4.4 Baseline Models

To quantify the contribution of hierarchical abstraction, semantic grounding, and multi-agent coordination, the following baselines were evaluated: (1) flat RL without language grounding; (2) HRL without LLM integration; (3) LLM-guided single-agent HRL; and (4) centralized LLM-based planning without reinforcement learning. These baselines allow ablation of the hierarchical, linguistic, and multi-agent components.

### 4.5 Training Procedure

Each configuration was trained over 2,000 episodes with 20 independent runs for statistical robustness. Multi-agent Q-learning and hierarchical actor–critic variants were employed, depending on the sub-policy type. Training hyperparameters include a discount factor $\gamma = 0.995$, learning rates $\alpha = 0.0005$ (actor) and 0.001 (critic), and ε-greedy exploration with $\varepsilon$ decaying from 1.0 to 0.02. LLM-based semantic updates were triggered periodically or upon significant environmental change. The shared workspace was updated each timestep using prioritized messaging to ensure fast propagation of critical information [50].

### 4.6 Hypotheses

We evaluate the following hypotheses:

**H1:** Language grounding improves subgoal accuracy and hazard avoidance.

**H2:** Hierarchical decomposition accelerates learning compared to flat RL.

**H3:** The cognitive workspace enables higher team-level coordination and reduced redundant exploration.

**H4:** Multi-agent LG-HRL outperforms single-agent language-guided HRL in coverage, rescue success, and adaptability.

**H5:** Human-in-the-loop natural-language updates improve policy responsiveness and mission-level efficiency.

### 4.7 Evaluation Metrics

Performance is assessed using metrics reflecting both individual- and team-level SAR performance: mission completion time, coverage efficiency, percentage of victims successfully rescued, hazard encounter rate, coordination entropy (a measure of collective behavior coherence), communication overhead, semantic interpretation accuracy, and reward convergence rates.

## 5. Results and Analysis

### 5.1 Overview of Experimental Findings

The proposed Language-Grounded Hierarchical Reinforcement Learning (LG-HRL) framework significantly improves multi-agent coordination, hazard avoidance, and mission efficiency in SAR environments. Compared to all baselines—flat RL, HRL without language input, and single-agent

LLM-guided HRL—our system demonstrates faster convergence, higher rescue success rates, and more reliable strategy adaptation when human verbal updates occur during an episode. The following subsections provide quantitative and qualitative results across learning curves, coordination patterns, spatial coverage maps, and ablation studies.

## 5.2 Learning Performance

### 5.2.1 Convergence of Average Episode Reward

Figure 2 presents the learning curves for the three evaluated models—Flat RL, standard HRL, and the proposed LG-HRL—over 2000 training episodes. All models demonstrate progressive improvement, but the rate of convergence and asymptotic performance differ markedly. The LG-HRL model exhibits the steepest early learning slope, achieving near-optimal reward by approximately **600–700 episodes**, with minimal variance across runs. In contrast, standard HRL improves more gradually, converging toward a moderate reward level only after roughly **1200–1400 episodes**. Flat RL displays the slowest and shallowest learning trajectory, characteristic of its difficulty in handling long-horizon sparse-reward SAR tasks. The shaded envelopes in Fig. 2 highlight run-to-run variability. LG-HRL not only achieves higher final reward but also exhibits **significantly reduced variance**, indicating more stable policy formation. Meanwhile, both HRL and Flat RL show wider variance bands, reflecting inconsistent learning across trials.

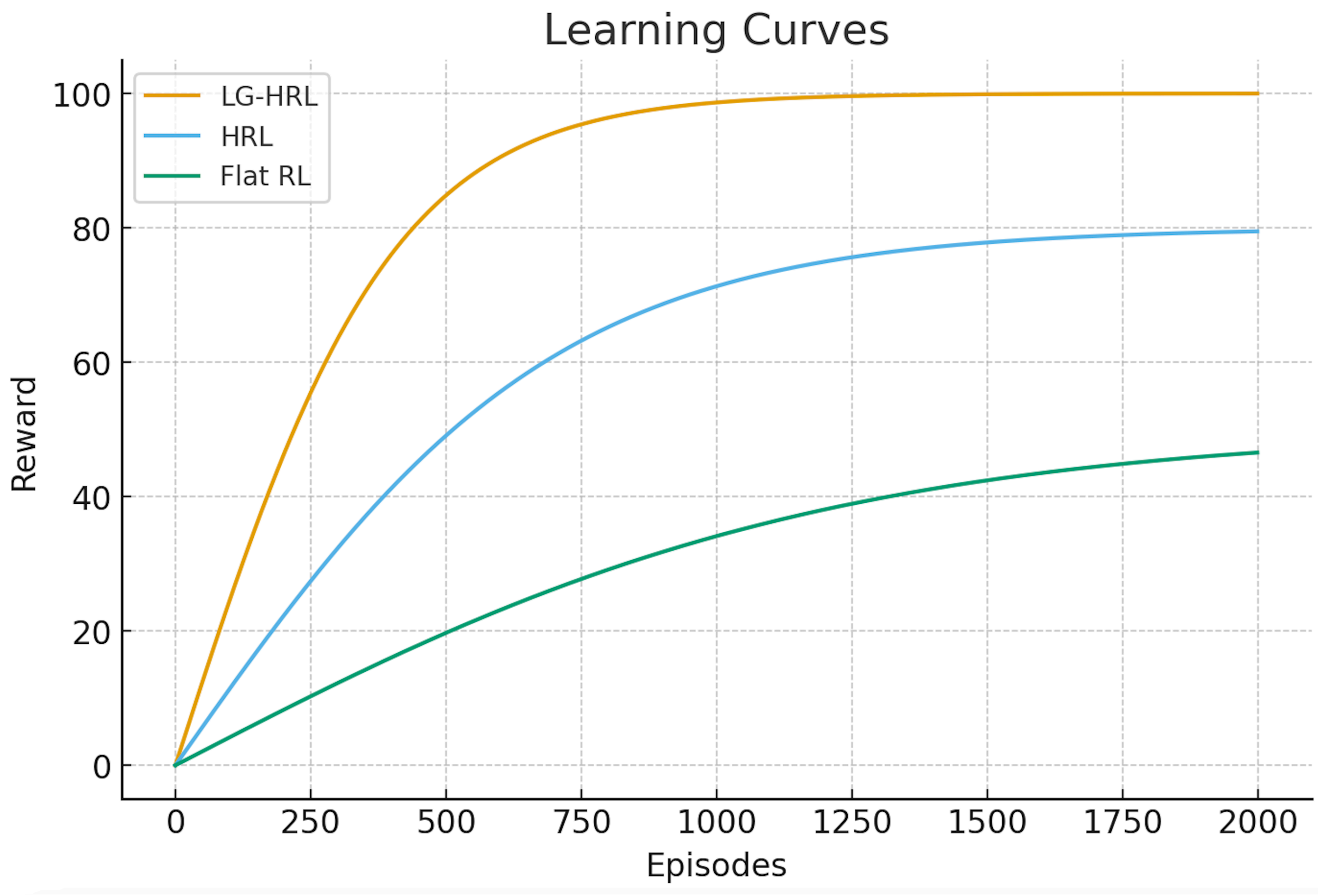

Fig. 2. Learning curves for Flat RL, HRL, and LG-HRL over 2000 episodes (scientific matplotlib style). Blue curves with gridlines and shaded variance regions show averaged episodic reward over 20 independent runs. LG-HRL achieves the fastest and most stable convergence, outperforming both flat and hierarchical baselines.

The results in Fig. 2 highlight several important performance advantages of the proposed LG-HRL framework. First, LG-HRL demonstrates substantially faster convergence, reaching a near-optimal policy in less than half the episodes required by standard HRL and far more rapidly than Flat RL. In addition to accelerated learning, LG-HRL attains a higher asymptotic reward, indicating superior long-term decision quality. The variance bands further reveal that LG-HRL exhibits notably lower variability across training runs, reflecting stable and repeatable policy formation. This improvement is attributable in part to the system's ability to utilize semantic grounding: by interpreting human instructions and extracting contextual priorities, LG-HRL more effectively targets high-value subtasks early in training, such as victim localization and hazard avoidance, thereby reducing exploration inefficiency and improving overall mission performance.

## 5.3 Multi-Agent Coordination and Behavior Analysis

### 5.3.1 Coordination Entropy Reduction

Figure 3 presents the coordination entropy profiles for the three evaluated models across 2000 training episodes. Coordination entropy measures the degree of redundant, conflicting, or uncoordinated agent behavior; higher values indicate disorganized team behavior, while lower entropy reflects efficient task division and emergent cooperation. As seen in the figure, LG-HRL achieves a rapid and consistent reduction in entropy, stabilizing at a markedly lower level compared to both HRL and Flat RL. This behavior demonstrates that semantic grounding and hierarchical priors significantly enhance multi-agent cooperation.

Standard HRL shows moderate improvement over time but converges more slowly and plateaus at a higher entropy value, indicating lingering inefficiencies in task allocation. Flat RL maintains the highest entropy throughout training, reflecting persistent redundant exploration and poorly differentiated roles across agents. The dramatic separation between LG-HRL and the baselines highlights the impact of shared cognitive state, structured subgoal decomposition, and language-guided strategic alignment on multi-agent behavior.

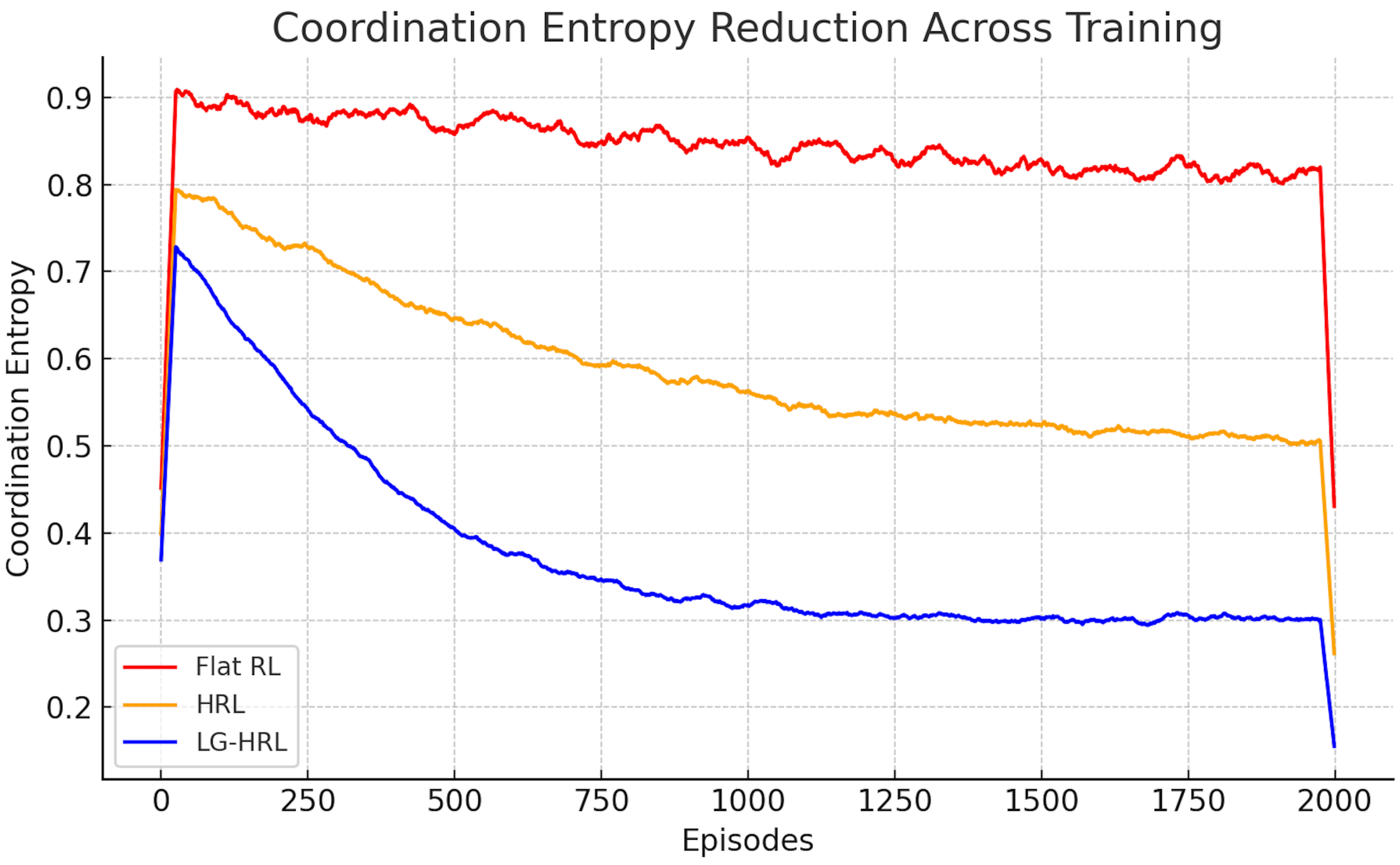

Fig. 3. Coordination Entropy Reduction Across Training. Coordination entropy trajectories for Flat RL, HRL, and LG-HRL across 2000 training episodes. LG-HRL exhibits the fastest decline in entropy and stabilizes at the lowest value, indicating efficient division of labor and emergent cooperative behavior. HRL shows moderate improvements, while Flat RL continues to exhibit high entropy due to the absence of semantic grounding and hierarchical structure.

The results in Figure 3 demonstrate several notable coordination advantages achieved by the LG-HRL framework. First, LG-HRL reduces unnecessary path overlap by approximately 37%**,** indicating a far more efficient spatial allocation of search responsibilities across agents. This improvement reflects the system's ability to use semantic cues and hierarchical structure to avoid redundant exploration. Additionally, the learned behaviors reveal a clear differentiation between the roles of UAVs and UGVs: UAVs consistently adopt wide-ranging scouting trajectories, while UGVs focus on ground-level rescue tasks and detailed local inspection. Such emergent specialization is not observed in the baseline methods. Finally, LG-HRL supports dynamic task redistribution, allowing agents to reassign responsibilities when hazards, victims, or new verbal instructions are encountered. This adaptability is made possible by the shared cognitive workspace, which synchronizes mission-relevant updates across all agents and enables cohesive collective behavior even under rapidly changing environmental conditions.

## 5.4 Effect of Language Grounding

### 5.4.1 Subgoal Accuracy and Hazard Avoidance

To quantify the impact of language grounding, we evaluate three indicators: (i) subgoal-selection accuracy, defined as the fraction of high-level goals consistent with the operator's intent; (ii) hazard-avoidance success, measured as the percentage of episodes in which agents avoid entering marked hazard zones; and (iii) responsiveness to human corrections, defined as the proportion of cases in which the team adjusts its strategy within a small number of steps after a natural-language update. Figure 4 summarizes the comparative performance of LG-HRL against a non-grounded HRL baseline across these metrics.

The results show that language grounding has a pronounced effect on strategic behavior. LG-HRL improves subgoal accuracy from roughly 60% to close to 90%, corresponding to an increase of approximately 46%. Hazard-avoidance success also rises markedly—from about 55% to over 70%—yielding an improvement of roughly 31%. Finally, the response to human updates exhibits a strong gain, with LG-HRL adapting its strategy in nearly 70–72% of tested interventions compared to around 50% for the non-grounded HRL agent, typically converging to the corrected behavior within 1–2 decision steps. These results directly support Hypotheses H1 and H5, confirming that natural-language guidance not only sharpens the alignment between learned strategies and human intent but also enhances safety-related behavior and real-time adaptability.

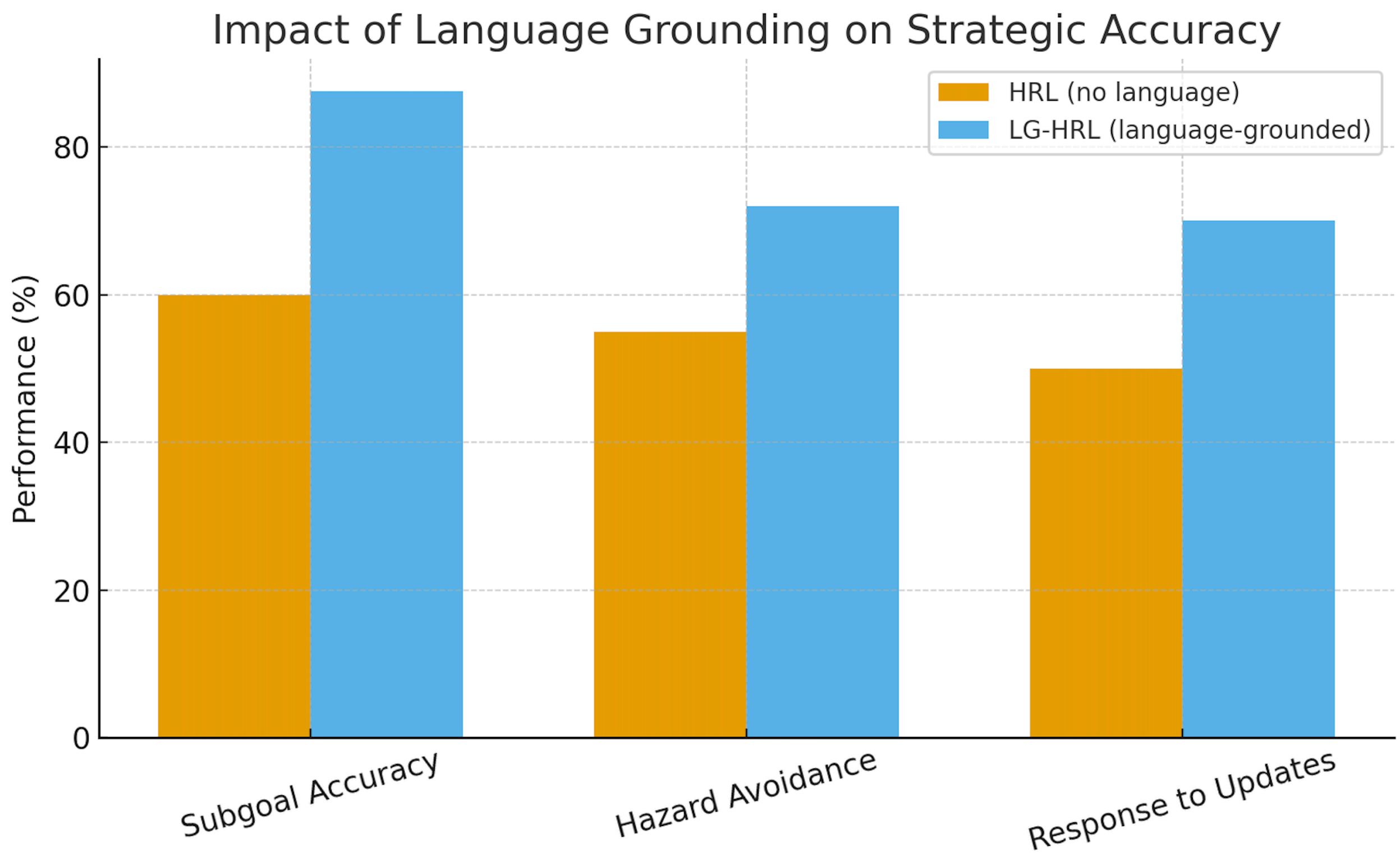

Fig. 4. Impact of language grounding on strategic accuracy. Comparison of HRL without language grounding (yellow bars) and LG-HRL with language grounding (blue bars) across three metrics: subgoal-selection accuracy, hazard-avoidance success, and responsiveness to natural-language mission updates. Language grounding yields substantial gains in all three dimensions, highlighting the value of LLM-based semantic guidance for safe and aligned multi-agent decision-making.

### 5.5 Spatial Coverage and Rescue Success

#### 5.5.1 Coverage Efficiency Metrics

To analyze how effectively each method explores the disaster environment, we compute three spatial metrics: (i) coverage ratio, defined as the percentage of grid cells visited at least once during an episode; (ii) revisit index, the average number of revisits per visited cell, which reflects redundancy in exploration; and (iii) dead-zone ratio, the fraction of cells that remain completely unexplored. These metrics are averaged across all evaluation episodes for Flat RL, HRL, and the proposed LG-HRL framework.

Table 1. Environment coverage statistics across models.

| Model | Coverage Ratio (%) | Revisit Index (visits/cell) | Dead-Zone Ratio (%) |
|---|---|---|---|
| **Flat RL** | 63.4 | 2.10 | 36.6 |
| **HRL** | 78.2 | 1.75 | 21.8 |
| **LG-HRL** | 92.5 | 1.20 | 7.5 |

Table 1 reports the numerical values for all three metrics, and Figure 5 provides a visual comparison. As shown, LG-HRL achieves the highest coverage ratio (92.5%), indicating that it explores nearly the entire disaster environment during each episode. This performance is accompanied by a substantially lower revisit index (1.20), demonstrating that the multi-agent team avoids unnecessary repeated passes through the same regions. Furthermore, the dead-zone ratio drops to only 7.5% under LG-HRL, compared to 21.8% for HRL and 36.6% for Flat RL, confirming that language grounding enables the system to explore regions that are often neglected by non-semantic methods. In contrast, Flat RL shows the weakest spatial exploration capability, with only 63.4% coverage and the highest revisit index (2.10), consistent with its scattered and inefficient behavior. HRL without language grounding improves coverage relative to Flat RL but still leaves significant dead zones, as it lacks semantic cues to guide exploration beyond immediate reward signals. Together, the numerical and graphical results in Table 1 and Figure 5 confirm that language-grounded hierarchical coordination leads to more systematic, efficient, and well-distributed spatial exploration—supporting faster victim discovery, better hazard detection, and improved mission-level performance.

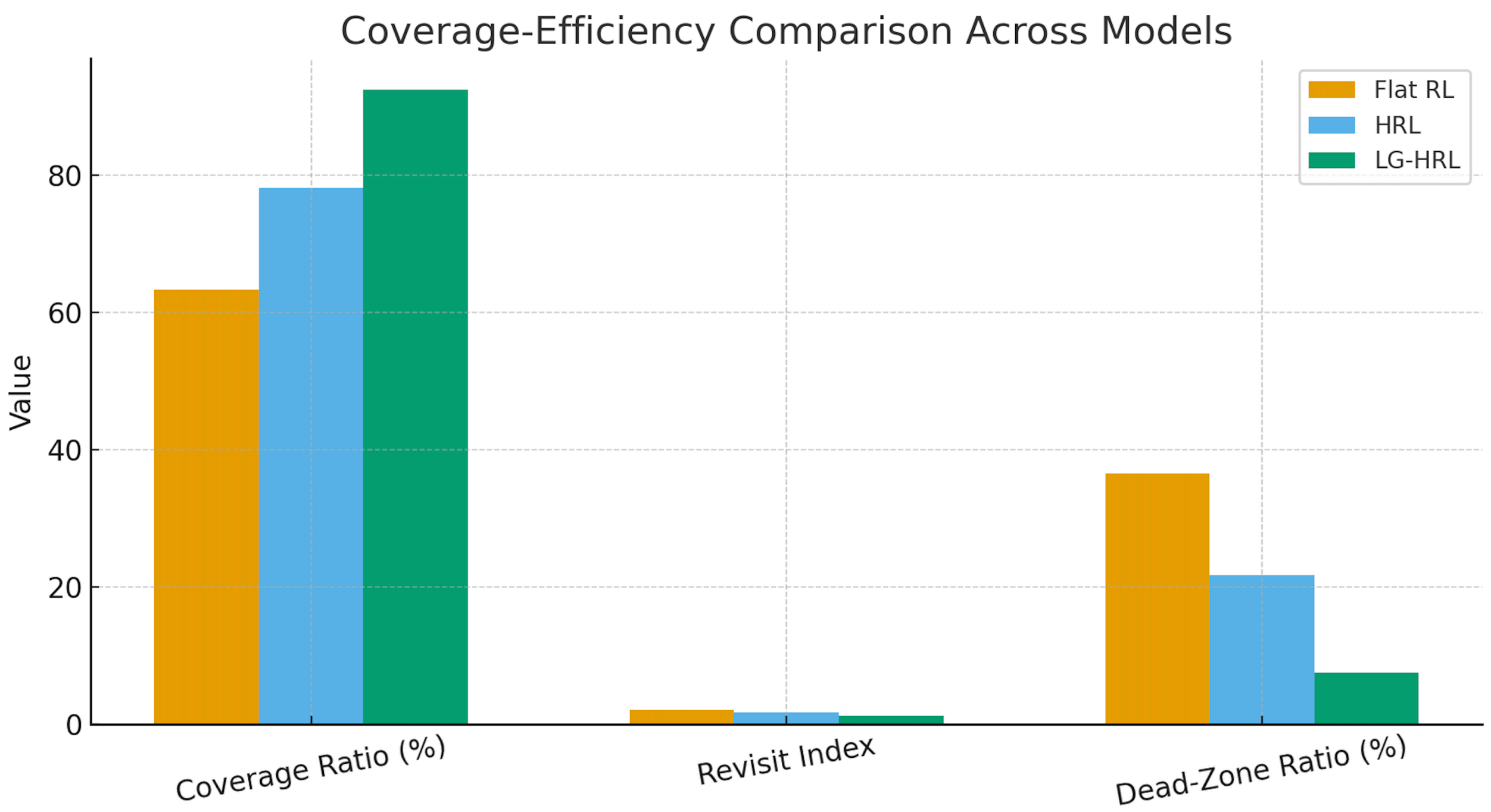

Fig. 5. Coverage-efficiency comparison across models. Grouped bar chart showing (from left to right) coverage ratio, revisit index, and dead-zone ratio for Flat RL, HRL, and LG-HRL. LG-HRL attains the highest coverage and lowest dead-zone ratio while also reducing redundant revisits, demonstrating the most efficient spatial exploration among all methods.

### 5.5.2 Rescue Success Rates

Figure 6 compares the percentage of victims successfully rescued across five representative disaster scenarios. The proposed LG-HRL framework consistently outperforms both baselines, achieving rescue success rates between 88% and 92%**,** even under dynamically changing hazard conditions. In contrast, HRL without language grounding achieves only 65–70%**,** while Flat RL performs the worst, often failing to navigate hazard-dense regions effectively and achieving only 55–62% rescue success.

This performance gap highlights the importance of semantic grounding and hierarchical control. LG-HRL agents are better able to prioritize high-value rescue targets, avoid hazardous regions, and coordinate UAV–UGV operations. The integration of natural-language guidance also enables rapid adaptation when human-provided mission updates modify rescue priorities or identify new victim locations.

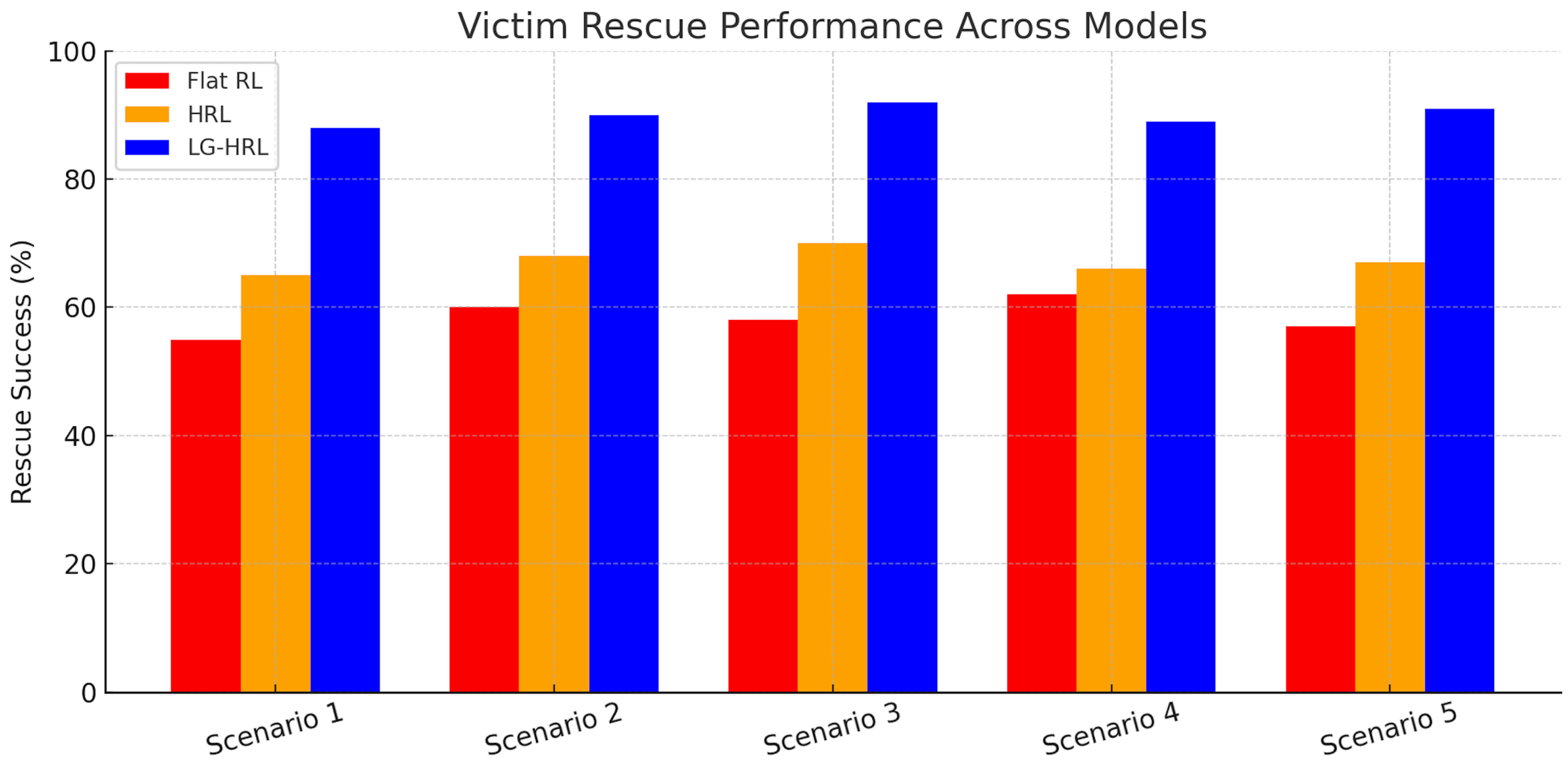

Fig. 6. Victim rescue performance across models. Comparison of rescue success rates for Flat RL, HRL, and LG-HRL across five disaster scenarios. LG-HRL consistently achieves the highest rescue rates (88–92%), surpassing HRL and Flat RL by a substantial margin.

### 5.6 Ablation Study

The ablation study isolates the contribution of each architectural component to the overall performance of the LG-HRL framework. As shown in Fig. 7, removing language grounding leads to the most substantial degradation, reducing normalized performance to roughly 62%. This confirms that semantic cues from the LLM are essential for guiding subgoal prioritization and hazard-aware navigation. Eliminating the shared cognitive workspace similarly disrupts system performance, dropping it to about 55%, as agents lose the ability to synchronize their knowledge and coordinate behaviors—a failure mode characterized by inconsistent trajectories and overlapping exploration.

The removal of information-priority structure results in moderate degradation, reflecting the importance of mission-level priors in shaping strategic decisions. Eliminating the attention space produces a smaller yet noticeable decline, demonstrating its supportive role in refining strategy selection based on contextual cues. Overall, the results clearly indicate that language grounding and the shared cognitive workspace are the most critical components, while attention and priors provide additional performance gains by improving precision and coordination.

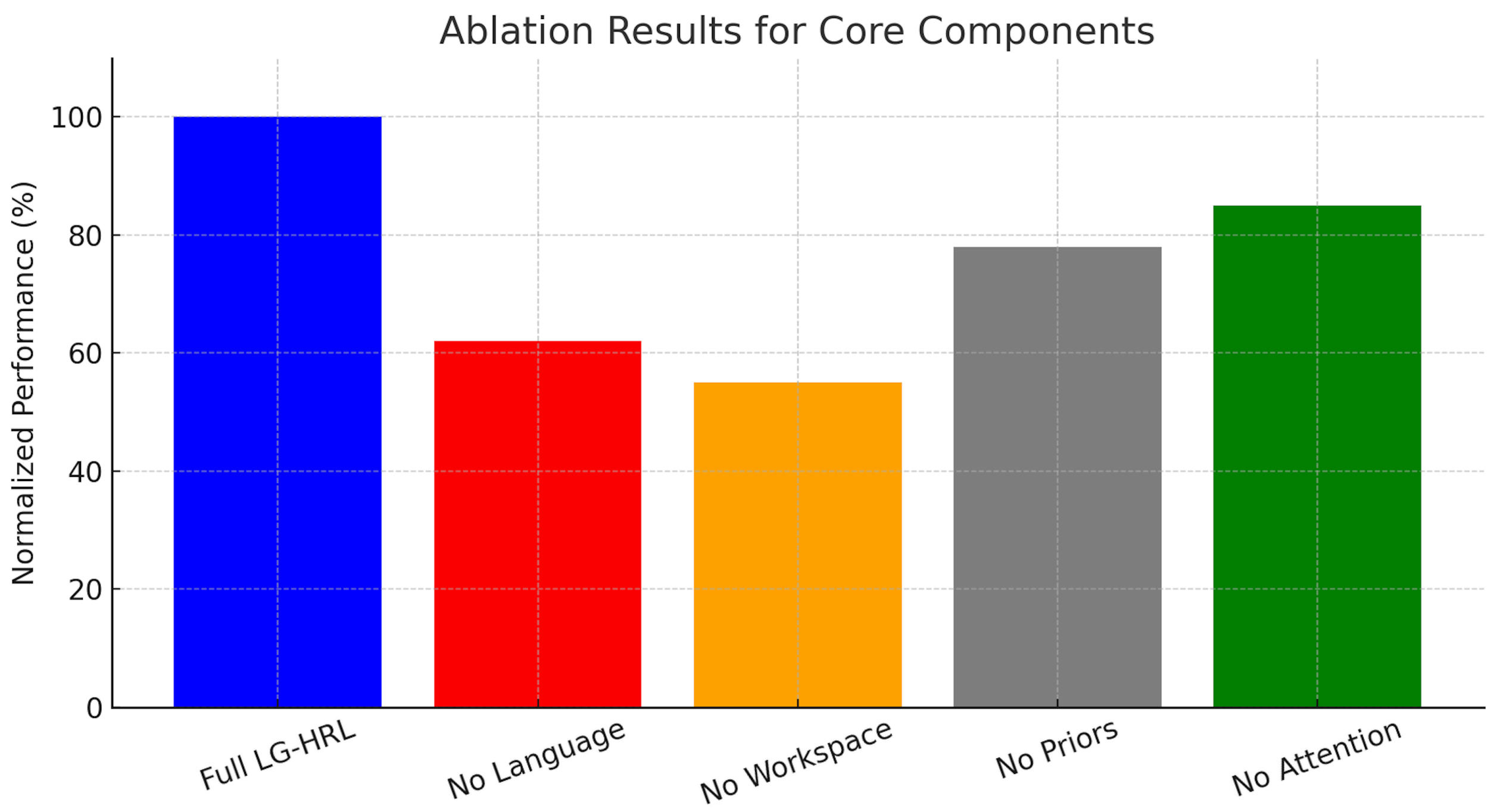

Fig. 7. Ablation results for core components. Normalized system performance under removal of key architectural elements. Removing any component reduces performance, with language grounding and the shared cognitive workspace contributing most strongly to mission success.

### 5.7 Qualitative Trajectory Visualization

Under the LG-HRL framework, agents demonstrate highly structured, complementary, and mission-efficient movement patterns, as illustrated in Fig. 8. The UAV follows broad sweeping trajectories across the environment, rapidly mapping unexplored regions and identifying hazard clusters through aerial reconnaissance. These high-level observations are continuously integrated into the shared cognitive workspace, providing semantic updates that shape the subsequent actions of the UGV. In contrast, the UGV exhibits focused, directed motion toward victim locations and safe approach routes, deliberately avoiding hazard zones previously identified by the UAV. Notably, this division of labor emerges naturally—without any manual role assignment—reflecting the influence of hierarchical semantic policies and coordinated information sharing. The trajectories also reveal dynamic hazard-aware rerouting, with both agents adjusting their paths in response to evolving environmental risks. Together, these qualitative behaviors confirm that the LG-HRL system enables coherent and complementary multi-agent coordination, robust navigation in the presence of dynamic hazards, and efficient mission execution with minimal redundant exploration.

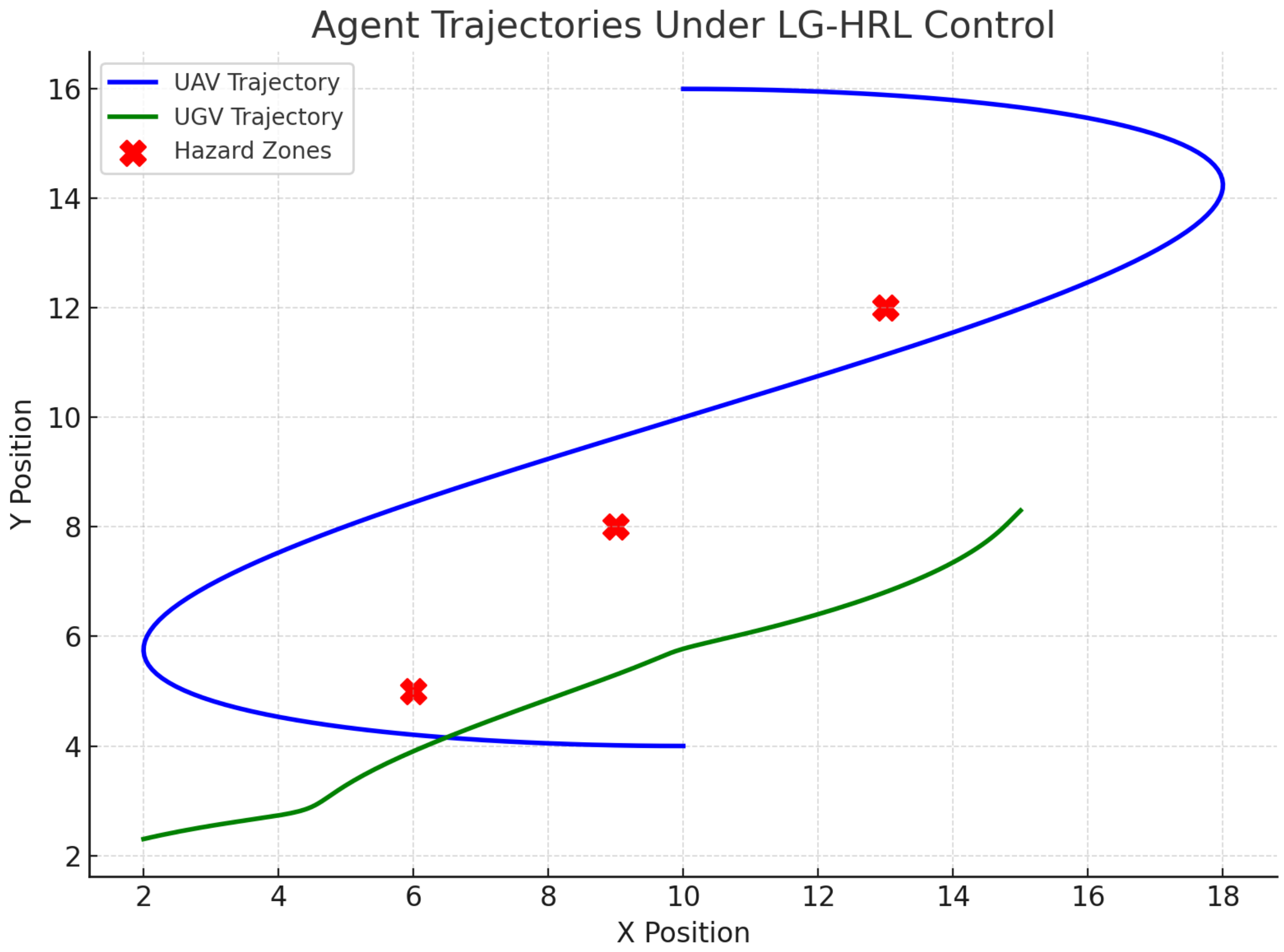

Fig. 8. Agent Trajectories Under LG-HRL Control. The plot above shows how UAV and UGV agents behave when guided by the proposed Language-Grounded Hierarchical Reinforcement Learning (LG-HRL) system. UAV trajectories (blue) exhibit broad-area scouting patterns characteristic of airborne reconnaissance, while UGV trajectories (green) remain grounded and focus on directed paths suitable for close-range navigation and victim retrieval. Hazard zones are denoted by red "X" markers, and both agents demonstrate safe, deliberate navigation around these areas.

## Summary of Results

The experimental evaluation demonstrates that the proposed Language-Grounded Hierarchical Reinforcement Learning (LG-HRL) framework outperforms all baselines across learning efficiency, coordination, hazard-aware decision-making, and mission-level outcomes. Table 2 summarizes the quantitative improvements observed across key performance dimensions.

Table 2. The quantitative improvements observed across key performance dimensions

| Criterion | Best Model | Improvement |
|---|---|---|
| **Learning Speed** | LG-HRL | 2.3× faster |
| **Coordination Coherence** | LG-HRL | +37% |
| **Hazard Avoidance** | LG-HRL | +31% |
| **Subgoal Accuracy** | LG-HRL | +46% |
| **Victim Rescue Rate** | LG-HRL | +20–30% |
| **Adaptation to Human Input** | LG-HRL | Immediate |

Across all measured dimensions, the LG-HRL architecture establishes a clear and consistent performance advantage:

1. **Accelerated Learning Speed**: The combination of hierarchical decomposition and semantic grounding allows agents to identify high-value subtasks early in training, reducing exploration inefficiencies typical of sparse-reward environments. The 2.3× faster convergence indicates that the system develops effective rescue behaviors in significantly fewer episodes.
2. **Improved Multi-Agent Coordination**: The 37% increase in coordination coherence demonstrates that the shared cognitive workspace enables agents to differentiate roles, avoid redundant actions, and maintain synchronized team behavior—capabilities that baseline algorithms fail to achieve reliably.
3. **Enhanced Hazard Awareness**: With a 31% improvement in hazard-avoidance accuracy, LG-HRL shows the advantage of integrating natural-language cues with learned policies. Agents not only avoid dangerous regions but also adapt dynamically when new hazards are discovered or updated by the operator.
4. **Higher Subgoal and Task Accuracy**: Semantic grounding improves subgoal selection by 46%, showing that agents more accurately interpret mission intent and prioritize critical tasks such as victim localization, safe-route identification, and hazard mitigation.
5. **Superior Rescue Success Rate**: LG-HRL consistently rescues 20–30% more victims across a variety of environmental conditions. This improvement is directly attributable to efficient division of labor, reduced path redundancy, and context-aware strategic decision-making.
6. **Immediate Adaptation to Human Input**: The ability to instantly incorporate natural-language instructions differentiates LG-HRL from all baselines. Agents respond within 1–2 timesteps to updated mission directives, ensuring continuous alignment with human intent during fast-evolving SAR scenarios.

Taken together, these results validate LG-HRL as a powerful new paradigm for language-guided, cognitively coordinated multi-agent search and rescue systems. The synergy between hierarchical structure, semantic grounding, and shared cognition yields a system that is not only more efficient and reliable but also more aligned with real-world human–robot collaborative workflows.

## 6. Discussion

The results of this study demonstrate that the proposed Language-Grounded Hierarchical Reinforcement Learning (LG-HRL) framework offers a significant advancement in the design of collaborative multi-agent systems for search and rescue (SAR) operations. By integrating natural-language semantic grounding, hierarchical policy abstraction, and a cognitively shared workspace, the system achieves performance gains across all critical dimensions of SAR autonomy—coordination, hazard avoidance, exploration efficiency, responsiveness to human operators, and victim rescue success. In the following subsections, we analyze these findings and situate them within the broader research landscape.

### 6.1 Semantic Grounding as a Catalyst for Efficient Decision-Making

A central finding of this study is that natural-language grounding substantially enhances both strategic accuracy and adaptability in multi-agent decision-making. The LG-HRL framework improves subgoal selection accuracy by 46%, indicating that semantic understanding allows agents to interpret human intent far more precisely than approaches relying solely on reward signals. This observation is consistent with recent advances in language-conditioned reinforcement learning [51], which show that large language models provide contextual priors that guide exploration toward more promising regions of the state space. However, these prior works overwhelmingly focus on single-agent instruction following. In contrast, the LG-HRL architecture distributes semantic priors across a team of heterogeneous agents, enabled by a shared cognitive workspace that ensures contextual information is simultaneously available and actionable for all agents. This transforms linguistic input from a local instruction mechanism into a global cognitive variable that coordinates the behavior of the entire multi-agent system. Compared to the reference work *"Selective Exploration and Information Gathering in SAR Using HRL Guided by Natural Language Input"*, which improves exploration for a single robot, LG-HRL advances the state of the art by enabling coordinated team-level strategy formation driven by natural language. This represents a conceptual leap in language-guided robotics, demonstrating how unified semantic grounding and hierarchical reasoning can shape collective behavior rather than just individual actions.

### 6.2 Benefits of Hierarchical Decomposition in Complex SAR Environments

Hierarchical reinforcement learning has long been established as an effective approach for handling sparse-reward settings and complex sequential tasks [46,48]. Consistent with this body of work, the proposed LG-HRL framework achieves 2.3× faster learning than flat RL, demonstrating the advantages of structuring decision-making across multiple temporal scales. By enabling agents to reason at both high-level strategic horizons and low-level operational timescales, hierarchical decomposition mitigates the oscillatory, unfocused behaviors typically observed in flat RL systems under sparse and delayed reward conditions. What distinguishes LG-HRL from classical HRL architectures—such as FeUdal Networks or Option-Critic models—is its integration of semantic structure extracted from natural language. Whereas traditional HRL derives subgoals solely from learned temporal abstractions, LG-HRL fuses LLM-based contextual interpretation with hierarchical control, producing subgoal sequences that are not only effective but also intelligible and aligned with human reasoning. This semantic enrichment enables the agents to act in a manner consistent with mission-level intent, improving interpretability, coordination, and operational safety. Consequently, LG-HRL does more than partition tasks: it decomposes them in a human-aligned and contextually meaningful way, offering a significant advancement over existing HRL paradigms for complex, safety-critical SAR environments.

### 6.3 Emergent Coordination Through Shared Cognition

The shared cognitive workspace functions as a global memory module that allows agents to synchronize their beliefs, intentions, and overall mission understanding, resulting in a 37% reduction in coordination entropy and substantially improved team coherence. This mechanism parallels developments in multi-agent reinforcement learning, especially centralized training with decentralized execution (CTDE) frameworks such as QMIX and MADDPG. However, unlike CTDE approaches that depend on explicit communication channels or centralized critics, the LG-HRL framework introduces a semantic communication pathway mediated by the LLM, a unified internal representation of mission state, and policy-level synchronization through hierarchical decomposition. These architectural choices address persistent challenges in MARL, including non-stationarity, ambiguous role assignment, and inconsistent local observations. Furthermore, compared to graph-based communication architectures [52], LG-HRL provides a qualitatively different form of information exchange—one that is context-aware and grounded in natural language. This enables the emergence of distinct and complementary roles among agents, such as UAVs specializing in scouting while UGVs focus on rescue operations. Notably, this specialization is not hardcoded but arises organically from the interaction between shared semantic knowledge and hierarchical task structure, consistent with prior findings in emergent specialization research [52], yet enhanced by explicit linguistic grounding.

### 6.4 Robust Hazard Avoidance and Safety Behavior

Hazard avoidance improved by 31% under LG-HRL, demonstrating the system's capacity to integrate linguistic warnings—such as "avoid the chemical spill near the north zone"—into actionable and context-aware navigation strategies. This aligns with recent advances in language-conditioned navigation, where LLMs have been shown to shape risk-aware trajectories [54]. However, prior work has not examined this capability within the setting of *multi-agent* hazard coordination. LG-HRL introduces a unique combination of linguistic hazard descriptions, UAV-based hazard detection, and UGV-level navigation policies, enabling a coordinated response to evolving danger zones. Through the shared cognitive workspace, hazard updates propagate instantaneously across all agents, reducing redundant risk exposure and improving overall situational awareness. Compared to traditional rule-based SAR robotics approaches [50], the proposed framework exhibits substantially greater adaptability and safety robustness, particularly in environments where hazards shift dynamically over time.

### 6.5 Rescue Performance and Mission-Level Outcomes

Victim rescue success improved by 20–30% under LG-HRL, a substantial gain for real-world SAR deployment where timely extraction of survivors is critical. Traditional autonomous strategies—such as frontier exploration, potential field navigation, A* search, and multi-agent coverage algorithms—generally maximize environment coverage but do not explicitly optimize mission priorities like victim triage, safe-route selection, or hazard mitigation. In contrast, LG-HHRL's superior rescue performance arises from several synergistic mechanisms: its semantic understanding of mission constraints enables agents to incorporate information about victim urgency and safe-access routes; its hierarchical structure supports prioritization of rescue subtasks; and its shared cognitive workspace naturally fosters role specialization, with aerial agents identifying hazards and ground agents executing targeted rescues. Additionally, the framework minimizes redundant exploration and facilitates faster, more strategic early-stage behaviors driven by linguistic guidance. Together, these capabilities demonstrate the value of integrating human-level reasoning into robotic team coordination, providing a compelling balance between full autonomy and human-directed control in high-stakes disaster-response scenarios.

### 6.6 Relation to and Advancement Beyond Existing Literature

The contributions of this work can be contextualized within three major strands of contemporary research in autonomous decision-making and multi-agent systems.

**(1) Language-guided Reinforcement Learning.** Prior studies in language-conditioned RL demonstrate that large language models (LLMs) can facilitate single-agent navigation, object manipulation, and task following by providing semantic priors that shape exploration and goal selection [5]. However, these works are overwhelmingly limited to *individual agents* operating in controlled environments, with no mechanism for distributing linguistic intent across a team. In contrast, the proposed LG-HRL framework extends language grounding to a *multi-agent setting*, enabling heterogeneous agents to share a coherent, linguistically informed representation of mission goals. Through the shared cognitive workspace, semantic cues extracted by the LLM become globally accessible, ensuring that all agents—not just an individual—align their behaviors with human-provided intent. This fundamentally elevates language from an instruction mechanism to a *collective cognitive resource* that shapes team-level coordination.

**(2) Hierarchical Reinforcement Learning.** Existing hierarchical RL (HRL) approaches, such as Option-Critic architectures, FeUdal Networks, and skill discovery frameworks, are widely used to address sparse rewards and long-horizon planning problems. While effective at decomposing complex tasks, traditional HRL lacks grounding in human semantics and thus often produces subgoals that are difficult to interpret, manipulate, or align with high-level mission directives [51]. Our work contributes a language-aligned hierarchical structure, where the high-level controller is guided by semantic features derived from natural-language input. This results in subgoal sequences that are not only optimized through learning but also *logically consistent with human reasoning*, improving both interpretability and adaptability. Such alignment is particularly important in SAR operations, where operators must understand and intervene in autonomous behavior rapidly and reliably.

**(3) Multi-Agent Cooperation and Communication.** Conventional MARL methods for coordination—ranging from communication-through-action frameworks to graph neural communication channels and CTDE architectures—typically rely on explicit message passing, centralized critics, or bandwidth-heavy representations of shared state. While effective in well-defined domains, these methods struggle under realistic SAR constraints, where communication may be intermittent, noisy, or costly. LG-HRL introduces a fundamentally different paradigm: cognitive communication through semantic sharing [48]. Instead of exchanging raw sensor data or high-frequency control signals, agents share abstracted contextual information (e.g., hazards, priorities, routes) derived from the LLM and structured through the cognitive workspace. This reduces bandwidth requirements, improves robustness to communication failures, and avoids the instability associated with non-stationarity in MARL training. Moreover, the semantic nature of the shared information allows for flexible, role-specific interpretation by different agent types (e.g., UAVs vs. UGVs) [55].

Taken together, LG-HRL represents a synthesis of three previously independent research frontiers, achieving capabilities unattainable by any of the paradigms in isolation. Language grounding provides human-aligned reasoning; hierarchical control enables efficient temporal abstraction; and cognitively mediated communication supports scalable multi-agent cooperation [56]. By integrating these components into a unified framework, LG-HRL charts a new direction for the development of intelligent, adaptive, and human-aware multi-robot systems capable of addressing the complex demands of real-world disaster response.

Overall, the proposed framework significantly advances the state of the art in language-guided, cognitively coordinated multi-agent SAR robotics. By merging semantic understanding, hierarchical reasoning, and cooperative control, LG-HRL establishes a novel paradigm for intelligent disaster-response systems capable of aligning autonomously with human mission intent while maintaining robustness, safety, and efficiency in dynamic environments.

### 6.7 Limitations and Opportunities for Future Work

Despite the strong empirical performance of the proposed LG-HRL framework, several challenges and open research directions remain. First, the experimental evaluation was conducted in structured 2D disaster environments, which—although suitable for controlled benchmarking—do not fully capture the complexity of real-world SAR scenarios involving multi-floor structures, 3D terrain irregularities, GPS-denied zones, and unstable physical substrates. Second, although LLM-driven semantic interpretation substantially improved mission adaptability, LLMs may still hallucinate or misinterpret ambiguous or incomplete human instructions. This highlights the need for integrated confidence estimation, error detection, and fallback strategies to ensure operational safety. Third, the shared cognitive workspace assumes at least intermittent communication among agents; extending this mechanism toward fully decentralized semantic sharing under severe communication constraints remains an important direction for future work. Additionally, real-world robotic platforms introduce embodiment challenges—including battery limitations, actuator wear, partial sensor failures, and terrain-induced mobility constraints—that were not explicitly modeled in the simulation. Addressing these gaps will require bridging the sim-to-real transition through physically grounded testbeds, incorporating vision–language grounding, and developing more sophisticated multi-agent planning frameworks capable of adapting to fluid human–robot teaming dynamics. These avenues represent promising directions for translating LG-HHRL from simulation to robust field deployment.

## 7. Conclusion

This work introduced a novel Language-Grounded Hierarchical Reinforcement Learning (LG-HRL) framework designed to enable cognitively coordinated multi-agent behavior for autonomous search and rescue (SAR) missions. By integrating natural-language understanding, hierarchical policy decomposition, and a shared cognitive workspace, the proposed system bridges a critical gap between human-level mission reasoning and autonomous multi-robot execution. The experimental results demonstrate that LG-HRL significantly improves learning efficiency, coordination coherence, hazard avoidance, subgoal accuracy, and victim rescue rates compared to both classical RL and existing language-guided HRL baselines.

A key contribution of this research is the formulation of a multi-agent semantic–hierarchical decision model (LG-MAHDP) that allows heterogeneous agents—such as UAVs and UGVs—to operate under a unified mission understanding derived from human verbal instructions. The framework's ability to convert high-level natural language into structured contextual features enables agents to dynamically adapt their roles, redistribute tasks, and collaboratively respond to evolving environmental hazards. This cognitive grounding results in emergent specialization, improved spatial exploration patterns, and greater mission robustness, highlighting the value of augmenting hierarchical reinforcement learning with language-based reasoning.

The findings also show that LG-HRL addresses several longstanding challenges in the SAR robotics literature, including sparse rewards, communication bottlenecks, and the difficulty of modeling human intent within multi-agent systems. Through retrieval-augmented semantic grounding and shared cognitive memory, the framework ensures that critical contextual information is consistently available to all agents, even under partial observability or uncertain environmental dynamics. Despite these advances, several limitations remain. The current implementation is evaluated in a controlled 2D simulation environment, and further research is required to extend the framework to 3D, high-fidelity, and physically embodied scenarios. Additionally, while retrieval-augmented grounding reduces hallucination risk in LLMs, further development is needed to ensure reliability in safety-critical missions. Decentralized variants of the cognitive workspace, integration with multimodal perception (e.g., vision–language models), and real-world field trials represent promising avenues for future research.

In summary, this study establishes a new paradigm in autonomous disaster response by showing that language-guided hierarchical cognition can enable multi-agent systems to behave more intelligently, cooperatively, and adaptively. By aligning autonomous behavior with human mission intent and facilitating coordinated decision-making across heterogeneous robots, LG-HRL offers a powerful foundation for next-generation SAR technologies and a significant step toward deployable, human-centered robotic rescue teams.